# Vision: A Deep Learning Approach to provide walking assistance to the visually impaired

Nikhil Thakurdesai
*Department of Computer Science*
*Indiana University Bloomington*
nthakurd@iu.edu

Anupam Tripathi
*Department of Artificial Intelligence*
*Northwestern University*
anupam.t@u.northwestern.edu

Dheeraj Butani
*Department of Computer Engineering*
*K.J. Somaiya College of Engineering*
dheeraj.butani@somaiya.edu

Smita Sankhe
*Department of Computer Engineering*
*K.J. Somaiya College of Engineering*
smitasankhe@somaiya.edu

*Abstract*— **Blind people face a lot of problems in their daily routines. They have to struggle a lot just to do their day-to-day chores. In this paper, we have proposed a system with the objective to help the visually impaired by providing audio aid guiding them to avoid obstacles, which will assist them to move in their surroundings. Object Detection using YOLO will help them detect the nearby objects and Depth Estimation using monocular vision will tell the approximate distance of the detected objects from the user. Despite a higher accuracy, stereo vision has many hardware constraints, which makes monocular vision the preferred choice for this application.**

## 1. INTRODUCTION

Globally, it is estimated that approximately 1.3 billion people live with some form of vision impairment, with the majority of them being over the age of 50. The visually impaired have to be dependent on others for guidance. Walking canes and guide dogs also provide limited assistance. Canes have a very small radius for which they can be used and they cannot help the user to tell what obstacle lies in front with much certainty. They are also not very useful in case of upper body or head level obstacles.

Object detection from a picture or a live video stream of the surroundings removes the need to come in contact with the obstacle in order to identify it. However, detecting objects in the surroundings is not enough unless the user can map their depths. With the name of the object and the distance of it from them, the visually impaired might find it much easier to move around. The best way to communicate all this information to the visually impaired would be through an audio output.

In this paper, we have proposed a system that will provide walking assistance to the visually impaired, by performing Object Detection using You Only Look Once (YOLO) [1] algorithm and Depth Estimation using Monocular vision. Unlike the algorithms that use sliding window of the image to localize the object within the image, the YOLO algorithm, as its name suggests, looks at the complete image and uses a single convolutional layer to predict the bounding boxes and their confidence levels.

From multiple captures of the same scene from different viewpoints, it is possible to estimate the depth of it. It is similar to the working mechanism of human eyes. Two eyes provide us with different viewpoints, which makes it easier for us to interpret the distances from the objects in the scene. Stereo vision uses the two viewpoints and maps them in 3-D to generate a disparity map. It requires a pair of camera-calibrated images, to generate the disparity map and then calculate the actual distance from it, but having two cameras to capture a pair of images makes the apparatus both bulky and costly. Monocular vision uses a single camera and uses deep learning to solve the stereo matching problem and generate disparity maps.

## 2. LITERATURE REVIEW

Convolutional Neural Networks (CNNs) are too slow and computationally very expensive for object detection problems. It was impossible to run CNNs on so many patches generated by sliding window detector. R-CNN [2] solves this problem by using an object proposal algorithm called Selective Search, which reduces the number of bounding boxes that are fed to the classifier to close to 2000 region proposals. Still, running CNN on 2000 region proposals generated by Selective search takes a lot of time. Girshick et al. published a second paper in 2015, entitled Fast R-CNN [3]. The Fast R-CNN algorithm made considerable improvements to the original R-CNN, namely increasing accuracy and reducing the time it took to perform a forward pass; however, the model still relied on an external region proposal algorithm. Both R-CNN and fast R-CNN used selective search to find out the region proposal, which made it slow. So, Shaoqing Renetal. [4] came up with an object detection algorithm that eliminates the selective search algorithm and lets the network learn the region proposals.

All the previously discussed techniques look at the image multiple times which increases the overall computational time. To help increase the speed of deep learning based object detectors, both Single Shot Detectors (SSDs)[5] and YOLO [6] use a one-stage detector strategy. These algorithms treat object detection as a regression problem, taking a given input image and simultaneously learning bounding box coordinates and corresponding class label probabilities. SSD eliminates proposal generation and is simpler to train. The accuracy of SSD is close to that of Faster-RCNN [4] and its speed is close to YOLO algorithm. Redmon et al. [1] proposed a technique for real-time object detection. It is the state-of-the-art technology used for object detection. The YOLO algorithm creates bounding boxes around the objects. This is done by scanning the entire image in one go, and hence the name "You Only Look Once". It then uses convolutional network and non-max suppression on the detected objects. YOLOv2 [7] is the second version of the YOLO with the objective of improving the accuracy significantly while making it faster. YOLO v3 [1] makes prediction at three scales. It predicts more number of bounding boxes per image and is also better at detecting smaller images.

Almalioglu, Yasin, et al. [8] proposed a deep monocular visual odometry and depth estimation method using Generative Adversarial Networks (GAN). Their architecture consists of a generator which generates disparity maps, a viewer construction which reconstructs the missing view, and a discriminator which will try to predict whether the generated depth map was real or fake. Their architecture out performs all the competing unsupervised and traditional baselines in terms of pose estimation. Chen, Richard, et al. [9] have also proposed an approach which uses GAN for depth estimation. They have paired the RGB image with their corresponding ground truth disparity map and trained the discriminator. Simultaneously, the generator also tries to generate disparity maps, which are then paired with the original RGB images and again sent to the discriminator. The loss is then used to train the generator to generate better and better disparity maps. Both these approaches achieve decent results, though there are certain areas GAN's need to be worked on such as mode collapse, non-convergence, diminished gradient, etc. Xie, Junyuanet al. [10] proposed a Deep3D architecture which would synthesize a disparity shifted right image from the left image. This architecture was used by Luo, Yue, et al. [11] to first synthesize a right view from the left image and then they have proposed a stereo matching network which predicts the disparity map using the left image and the synthesized right image.

## 3. EXPERIMENTAL SETUP

### 3.1. Database

The model used for object detection by us was trained on the COCO (Common Objects in Context) dataset [12], shown in Fig. 1(a), which contains images taken from everyday scenes. The dataset consists of 80 object categories of labeled and segmented images.

In order to train our model to generate disparity maps, we have used the KITTI dataset, shown in Fig. 1(b). The KITTI dataset [13] consists of image sequences taken from both the left and right cameras which are fixed at a constant distance throughout. The images consist of roads and naturally includes the vehicles and the people.

As an alternative to the internet dependent text-to-speech modules, we have created database of pre-converted audio files for all the text that can be outputted by the system.

### 3.2. Data preprocessing

#### 3.2.1. Mean Subtraction

Mean subtraction is used to help combat illumination changes in the input images in our dataset. We can therefore view mean subtraction as a technique used to aid our Convolutional Neural Networks. We first compute the average pixel intensity across all images in the training set for each of the Red, Green, and Blue channels. This implies that we end up with three variables: µR, µG, and µB. We then subtract the mean, µ, from each input channel of the input

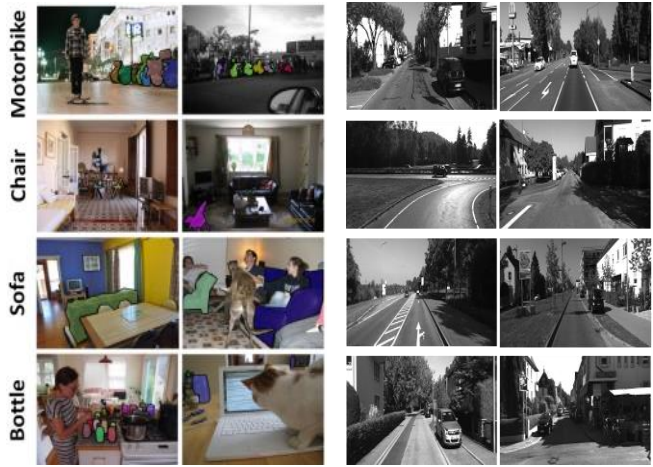

Fig. 1. (a) COCO DATASET (b) KITTI DATASET

image as shown in (1), (2) and (3).

$$R = R - \mu_R \quad (1)$$

$$G = G - \mu_G \quad (2)$$

$$B = B - \mu_B \quad (3)$$

*3.2.2. Scaling*

We may also have a scaling factor, which adds in a normalization:

$$R = \frac{R - \mu_R}{\sigma} \quad (4)$$

$$G = \frac{G - \mu_G}{\sigma} \quad (5)$$

$$B = \frac{B - \mu_B}{\sigma} \quad (6)$$

The value of sigma can be the standard deviation across the training set or can be set manually. We have chosen it to be 255 to normalize the data.

There was no preprocessing performed for the depth estimation module. Normalization of the data actually performed badly as compared to raw data.

4. METHODOLOGY

The proposed system consists of 3 modules; object detection, depth estimation, and text to speech. Fig. 2 gives an overview of the system architecture. The camera will capture photos, which will be sent to the object detection and depth estimation modules simultaneously. The detected objects and their corresponding depths will be constructed into a sentence and fed to the text to speech module. The output generated will be an audio file, which will guide the user about the nearby obstacles. The information about the distant obstacles will not be conveyed to the user.

*4.1. Object Detection*

A conventional convolutional neural network is not suitable for this problem as the length of the output layer will be variable because of the possibility of several objects belonging to the same class being detected and taking regions of interests could be computationally very expensive. Therefore, for object detection, algorithms like YOLO, SSD are preferred.

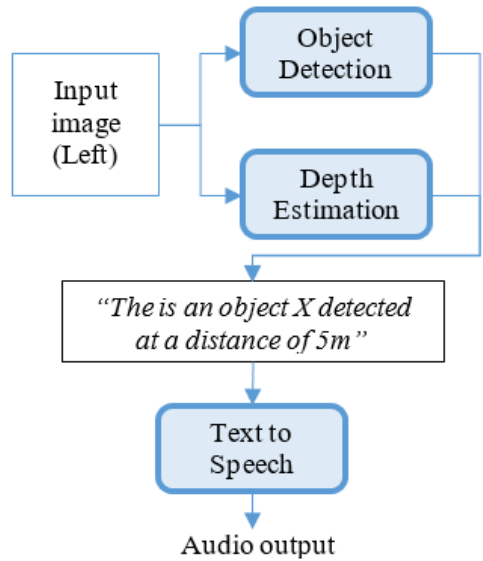

Fig. 2. SYSTEM ARCHITECTURE

For the YOLO algorithm, the image is divided into a SxS grid. If the center of any object falls into a grid cell, that grid cell is responsible for detecting that particular object. Each grid cell predicts B bounding boxes of different shapes and sizes. A detected object will be associated with the bounding box with which it has the greatest IoU (Intersection over Union). Each bounding box also has a confidence level associated with it.

Each bounding can be described using 5 descriptors: center of bounding box (bx, by), box height (bh), box width (bw) and the bounding box confidence level (pc) which is the probability of the object being in the box. Apart from these, each bounding box also has C class probabilities, one for each class of objects to be detected. With B such bounding boxes, each grid cell has B x (5+C) descriptors. To get bx, by, bw and bh from the network output, where tx, ty, tw, th is what the network outputs, cx and cy are the top-left co-ordinates of the grid and, pw and ph are anchors dimensions for the box. Since we are running our center coordinates prediction through a sigmoid function, the value of the output will be between 0 and 1. All four of these are relative to the top left corner of the grid cell which is predicting the object and normalized by the dimensions of the cell from the feature map. While detecting the objects, we also will be outputting the relative position of the object with respect to the cameras. With the coordinates of the bounding boxes, it is also possible to tell the position of the detected objects; whether they are on the right, left or to the front of the user.

Fig. 3 shows an example of the boundary box prediction. The image is divided into a 13x13 grid with each grid predicting 3 boundary boxes. The feature map has a depth of 3 x (5+80), i.e. 255 for the 80 classes. Each grid cell gives B bounding boxes, so the entire image will have S x S x B bounding boxes. Some of these boxes will have very low probability predictions, which can be eliminated straight away. The remaining boxes are passed through a "non-max suppression" that will eliminate possible duplicate objects and thus only leave the most exact of them.

YOLO v3 makes prediction across 3 different scales. The detection layer is used make detection at feature maps of three different sizes, having strides 32, 16, 8 respectively. The network downsamples the input image until the first detection layer, where a detection is made using feature maps of a layer with stride 32. Further, layers are upsampled by a factor of 2 and concatenated with feature maps of a previous layers having identical feature map sizes. Another detection is now made at layer with stride 16. The same upsampling procedure is repeated, and a final detection is made at the layer of stride 8. Therefore, for an input image of the same size, YOLOv3 predicts more boxes as compared to YOLOv2. For instance, YOLOv2 predicts 5 bounding boxes for each grid and with a grid size of 13 x 13, it will predict 845 boxes. Whereas YOLOv3 predicts 3 bounding boxes for each stride. With 507 anchor boxes for stride 32, 2,028 anchor boxes for stride 16 and 8,112 anchor boxes for stride 8, it predicts a total of 10,647 anchor boxes. This helps YOLO v3 get better at detecting small objects, a frequent complaint with the earlier versions of YOLO.

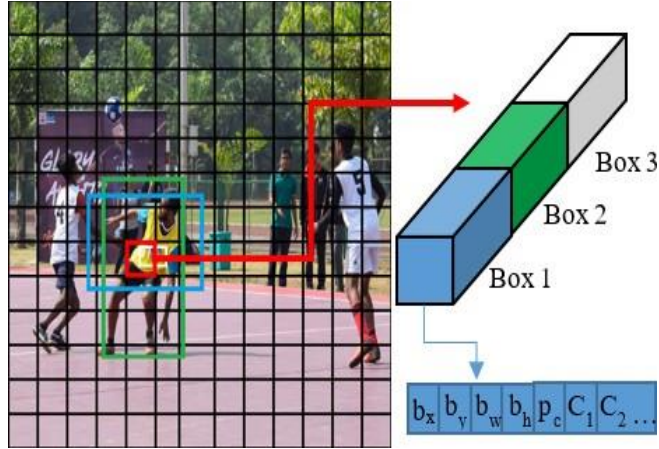

Fig. 3. EXTRACTING ANCHOR BOX ATTRIBUTES

This trained model will detect all the 80 classes in COCO dataset. Not all of these objects would be an obstacle for a visually impaired person at all times. For example, a user might not find a book lying outdoors as an obstacle. Similarly, if the user is indoor, the chances of finding a car as an obstacle would also be very low. Therefore, to save computational resources and time, and also for the convenience of the user, we have given the options of two modes for object detection: indoor and outdoor.

### 4.2. Depth Estimation

To get the distance between the user and a detected object, we need at least two viewpoints to map the image in 3-D and generate a depth map. Stereo vision uses two camera-calibrated images to perceive the three-dimensional structure of the world and generate its depth map. Calibration of the two cameras is done in order to know its intrinsic parameters such as the focal length, skew, distortion, image center etc. and its extrinsic parameters, which would describe its position and orientation in its surroundings.

To estimate the depth using stereo vision, we need the focal lengths and the distance between the two cameras. Fig. 4 shows the geometry behind the approach. Here, $x^l$ and $x^r$ are the distance between the optical axes and the projection of the object 'P' on the virtual planes of the two cameras..

If b is the baseline distance between the two cameras and f is the focal length of both the cameras, the perpendicular distance between object 'P' and the baseline of the cameras is given by:

$$Z = \frac{b * F}{x^l + x^r} \qquad (7)$$

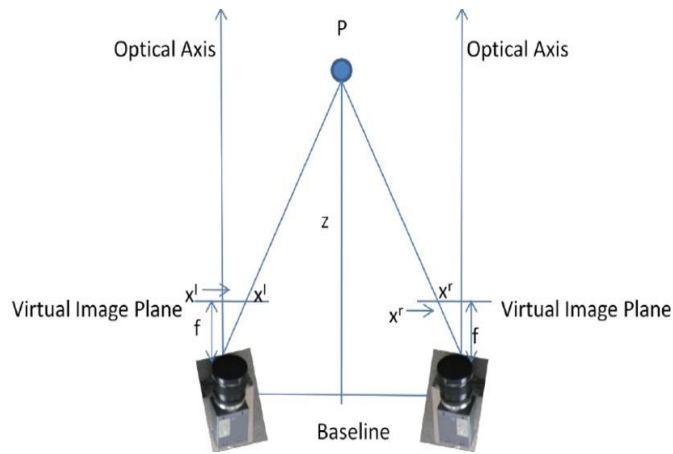

Fig. 4. STEREO VISION

To have two camera-calibrated images of the same view needs specialized stereo cameras. As we need at least two images from different viewpoints to estimate the depth, in this approach we try to generate the second image of the stereo pair of images using the first one. Fig. 5 shows an overview of the entire process. The right image is generated using the left image and then passed to stereo matcher to generate the disparity map. The architecture of the network used to generate the right image is shown in Fig. 6. The network consists of a series of convolution and de-convolution layers. The input to the "Disparity Introducing Layer" is a 300 x 300 x 33 matrix. Each of these 33 channels undergoes an element wise multiplication with the corresponding disparity shifted views of the original image as shown in the figure. The output of this layer is obtained by an element wise addition of all of these 33 products, which then undergoes a couple of convolutional layers to give the synthesized right image. The activation function used was Exponential Linear Unit (ELU) which is given in (8). ELU converged faster than Rectified Linear Unit (ReLU) and led to better results. L1 loss was used to train the network. A learning rate of 0.0003 was used to train this network. A weight decay of 1e-6 was used. The model was trained for 50 epochs. The batch size used was 16.

$$R(z) = \begin{cases} z & z > 0 \\ \alpha * (e^z - 1) & z \leq 0 \end{cases} \quad (8)$$

The generated right image from this network is passed to the stereo matcher network to generate the disparity map, architecture of which is shown in Fig. 7. Both the left and the generated right image were first passed through a couple of convolution layers for easier feature extraction. We then took a 1D correlation between the outputs of these convolutional layers. 1D correlation is an important step in the depth estimation of a scene. It provides us with valuable information in finding the corresponding pixel in the right image for a given pixel in the left image. Finally, after a series of convolution and deconvolution operations, we are able to generate the required disparity map. Again, ELU was used as the activation function. In this network, we observed experimentally, that the network was not able to converge well using the L1 loss function. Thus, we used the L2 loss function for this network. A learning rate of 0.0003, along with a weight decay of 1e-6 was used. The model was trained for 300 epochs, with a batch size of 16.

5. RESULTS

The output of the system is in the form of text, which is converted to audio. Fig. 8 shows the bounding box plotted for an input image.

In order to decide on the best object detection algorithm, we implemented and compared the results of a few object detection algorithms as shown in table 1. All these algorithms were implemented on several images. The table shows the confidence levels for one of those images. Table 1 shows the comparison between YOLO and a few other object detection algorithms, used on the same image. YOLO gives the best accuracy no matter the object.

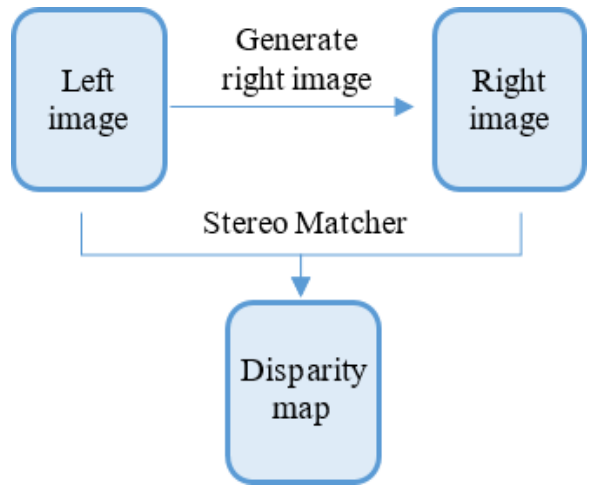

Fig. 5. OVERVIEW OF DEPTH ESTIMATION

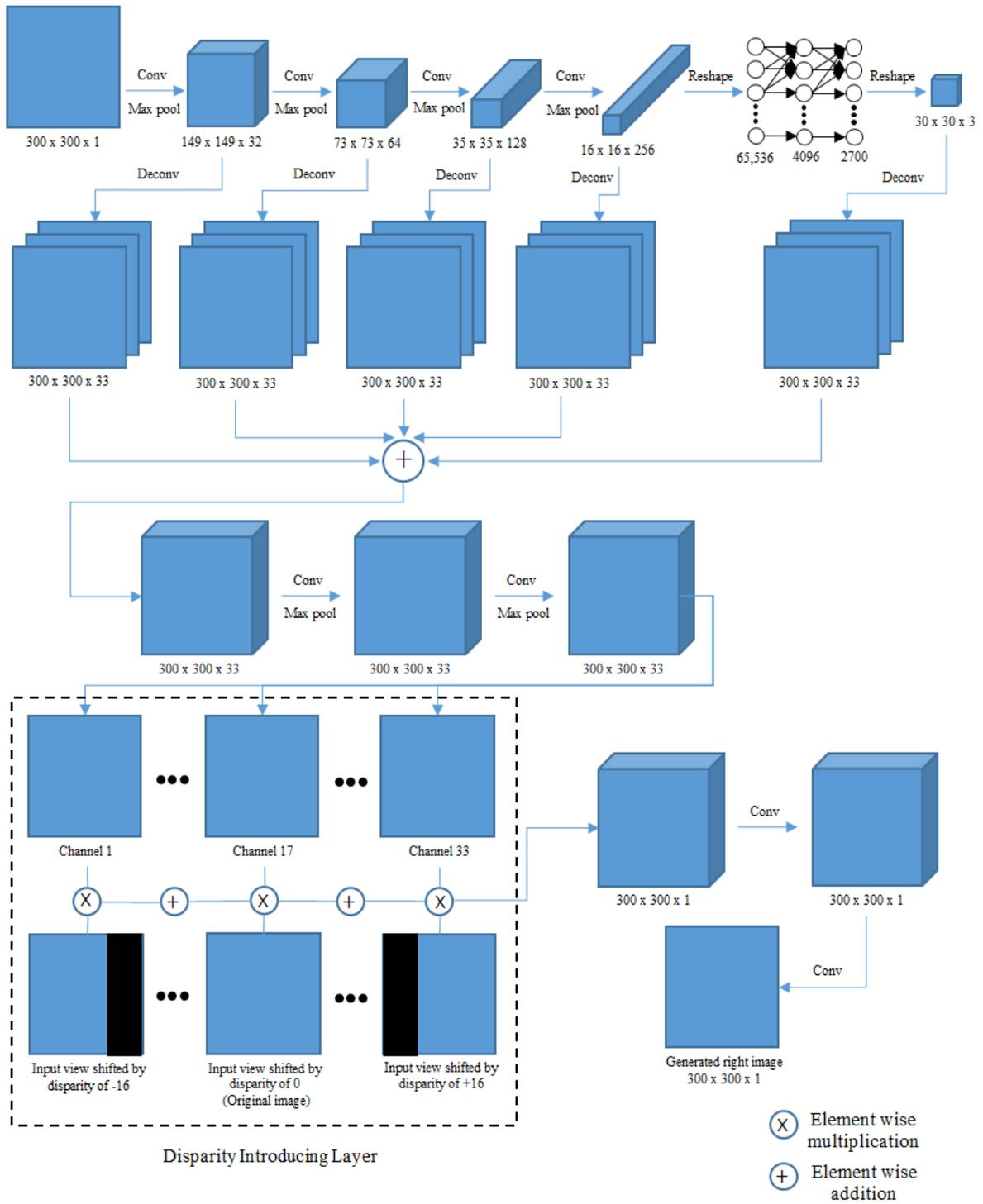

Fig. 6. ARCHITECTURE FOR SYNTHESIZING RIGHT IMAGE

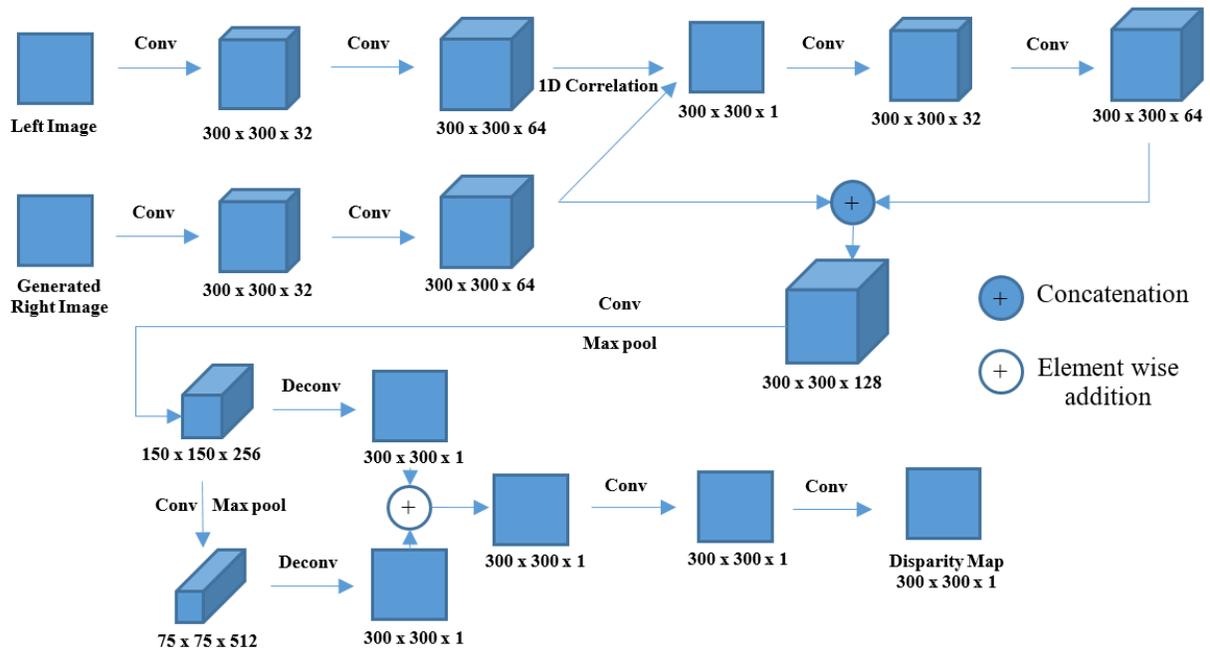

Fig. 7. STEREO MATCHING NETWORK

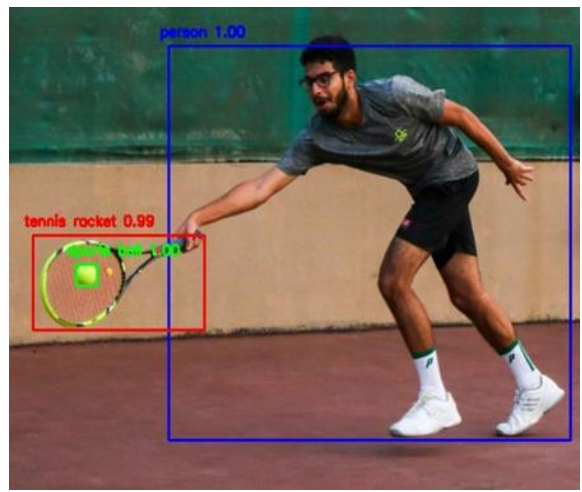

Fig. 8. OBJECT DETECTION OUTPUT

TABLE I  COMPARISON OF DIFFERENT OBJECT DETECTION MODELS

| Model | Confidence for object 1 | Confidence for object 2 |
|---|---|---|
| YOLO | 100% | 99% |
| Faster RCNN | 99% | 97% |
| SSD | 92% | 93% |

Fig. 9 (a) and 9 (b) show left image and the synthesized right image. Fig. 10 shows the disparity map for these two images, as generated by our network.

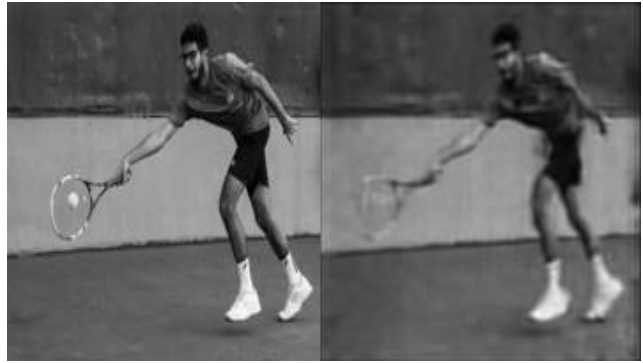

Fig. 9.  (a) LEFT (INPUT) IMAGE (b) SYNTHESIZED RIGHT IMAGE

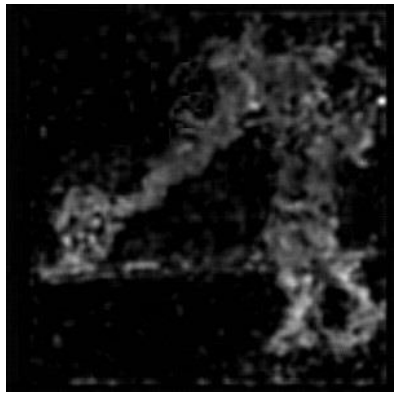

Fig. 10. GENERATED DISPARITY MAP

6. FUTURE WORK

The stereo matching network can be made more accurate in the future by training it for more epochs and a greater number of images from different datasets. Due to hardware limitations, we could only train the network on 586 images which is pretty less for most of the computer vision tasks. Still, we have managed to achieve decent results which can be useful as far as a visually impaired person is concerned. The network is giving decent accuracy for objects close to the camera. Using this, we can only tell the user about the objects that are very near to him/her, i.e. the objects that concern the user. We can also make an app and connect it to a remote server such as Amazon Web Server, Google Cloud Platform, etc. which will make it even easier for the user.